\documentclass{article}

\PassOptionsToPackage{numbers}{natbib}
\usepackage[preprint,position]{taylorgeospatial}

\usepackage[utf8]{inputenc}
\usepackage[T1]{fontenc}
\usepackage{xcolor}
\definecolor{linkblue}{HTML}{1B4F72}
\usepackage[
  colorlinks=true,
  linkcolor=linkblue,
  citecolor=linkblue,
  urlcolor=linkblue
]{hyperref}
\usepackage{url}
\usepackage{booktabs}
\usepackage{amsfonts}
\usepackage{amssymb}
\usepackage{amsmath}
\usepackage{nicefrac}
\usepackage{microtype}
\usepackage{graphicx}
\usepackage{subcaption}
\usepackage{wrapfig}
\usepackage{enumitem}
\usepackage[most]{tcolorbox}
\usepackage{longtable}
\usepackage{pdflscape}
\usepackage{needspace}
\usepackage{tabularx}
\tcbuselibrary{skins,breakable}

\definecolor{posblue}{HTML}{3B1E1C}
\definecolor{posbg}{HTML}{F4F4EB}
\definecolor{recred}{HTML}{FF4F2C}
\definecolor{recbg}{HTML}{FDE4DC}
\definecolor{checkblue}{HTML}{1B4F72}
\definecolor{checkbg}{HTML}{EAF2F8}
\newcommand{\rerunmark}{\ensuremath{\circlearrowright}}
\newcommand{\copiedmark}{\raisebox{-0.2ex}{\tikz[scale=1.35]{\draw[line width=0.35pt] (0,0.02) rectangle (0.13,0.15); \draw[line width=0.35pt,fill=white] (0.05,0) rectangle (0.18,0.13);}}}
\newcommand{\checkedbox}{\raisebox{-0.2ex}{\tikz[scale=0.24]{\draw[rounded corners=0.8pt,line width=0.55pt,draw=checkblue,fill=white] (0,0) rectangle (1,1); \draw[line width=0.8pt,draw=recred,line cap=round,line join=round] (0.22,0.52) -- (0.42,0.28) -- (0.78,0.74);}}}

\newtcolorbox{positionbox}[1][Position]{
  enhanced,
  colback=posbg,
  colframe=posblue,
  coltitle=white,
  title={\textbf{#1}},
  fonttitle=\bfseries\small,
  boxrule=0.4pt,
  arc=1pt,
  left=4pt,right=4pt,top=3pt,bottom=3pt,
  before skip=4pt, after skip=4pt,
}

\newtcolorbox{recbox}[1][Recommendation]{
  enhanced,
  colback=recbg,
  colframe=recred,
  coltitle=white,
  title={\textbf{#1}},
  fonttitle=\bfseries\small,
  boxrule=0.4pt,
  arc=1pt,
  left=4pt,right=4pt,top=3pt,bottom=3pt,
  breakable,
}

\newtcolorbox{checklistbox}[1][Reviewer checklist]{
  enhanced,
  colback=checkbg,
  colframe=checkblue,
  coltitle=white,
  title={\textbf{#1}},
  fonttitle=\bfseries\small,
  boxrule=0.4pt,
  arc=1pt,
  left=4pt,right=4pt,top=3pt,bottom=3pt,
  breakable,
}

\title{No One Knows the State of the Art\\in Geospatial Foundation Models}

\author{%
\begin{tabular}{c}
\textbf{Isaac Corley}\textsuperscript{1}\thanks{\texttt{Corresponding author: isaac.corley@taylorgeospatial.org}} \quad
\textbf{Nils Lehmann}\textsuperscript{2} \quad
\textbf{Caleb Robinson}\textsuperscript{3} \quad
\textbf{Gabriel Tseng}\textsuperscript{4} \quad
\textbf{Anthony Fuller}\textsuperscript{5,6}\\
\textbf{Hamed Alemohammad}\textsuperscript{7} \quad
\textbf{Evan Shelhamer}\textsuperscript{5,8} \quad
\textbf{Jennifer Marcus}\textsuperscript{1} \quad
\textbf{Hannah Kerner}\textsuperscript{1,9}\\[0.5em]
\normalfont
\textsuperscript{1}Taylor Geospatial \quad
\textsuperscript{2}Technical University of Munich \quad
\textsuperscript{3}Microsoft AI for Good Research Lab \\
\normalfont
\textsuperscript{4}Allen Institute for AI \quad
\textsuperscript{5}Vector Institute \quad
\textsuperscript{6}Carleton University \quad
\textsuperscript{7}Clark University\\
\normalfont
\textsuperscript{8}University of British Columbia \quad
\textsuperscript{9}Arizona State University\\[0.75em]
\normalfont\small
\href{https://research.taylorgeospatial.org/state-of-gfms}{\textcolor{recred}{\texttt{research.taylorgeospatial.org/state-of-gfms}}}
\end{tabular}
}

\begin{document}
\maketitle

\begin{abstract}
Geospatial foundation models (GFMs) have been proposed as generalizable backbones for disaster response, land-cover mapping, food-security monitoring, and other high-stakes Earth-observation tasks. Yet the published work about these models does not give reviewers or users enough information to tell which model fits a given task. \textbf{We argue that nobody knows what the current state of the art is in geospatial foundation models.} The methods may be useful, but the GFM literature does not standardize evaluations, training and testing protocols, released weights, or pretraining controls well enough for anyone to compare or rank them. In a 152-paper audit, we find 47 cross-paper disagreements of at least 5 percentage points, including 18 of at least 10 points, within model, benchmark, and protocol groups; 94/126 papers with pretraining data use a configuration no other paper uses; and 39\% of GFM papers release no model weights. This lack of community standards can be solved. We propose six concrete expectations: named-license weight release, shared core evaluations, copied-versus-rerun baseline annotations, variance reporting, one shared evaluation harness, and data-vs-architecture-vs-algorithm controls. These gaps are a coordination failure, not a fault of any individual lab; the authors of this paper, like many others in the GFM community, have contributed to them. Rather than just critiquing the community, we aim to provide concrete steps toward a shared understanding of how to evaluate and improve GFMs.

\end{abstract}

\section{Introduction}
\label{sec:intro}

The promise of geospatial foundation models is cheap, easy reuse across domains. A single pretrained Earth-observation backbone should transfer across sensors, geographies, label regimes, and downstream tasks: crop mapping, flood mapping, building extraction, forest monitoring, land-cover change, and more. That promise makes evaluation harder than ordinary model comparison. A paper may compare $M$ models across $N$ benchmarks, but the benchmarks differ in spatial resolution, modality, class definitions, geographic coverage, label quality, and whether the reported metric measures accuracy on small image patches or the quality of an actual map a user would rely on. \emph{This mirrors the benchmark-lottery problem} described in broader ML evaluation work~\citep{dehghani2021benchmarks}: benchmark choice can dominate apparent progress when communities lack shared protocols. The GFM community therefore needs \emph{clearer standards on how to test and compare GFMs}.

\citet{bommasani2021opportunities} introduced the term \emph{foundation model} for models trained on broad data that can be adapted to many downstream tasks; BERT~\citep{devlin2019bert}, GPT-3~\citep{brown2020language}, CLIP~\citep{radford2021learning}, DINO~\citep{caron2021emerging}, SAM~\citep{kirillov2023segment}, ImageNet-pretrained~\citep{deng2009imagenet} ResNets~\citep{he2016deep}, and ViTs~\citep{dosovitskiy2020image} became useful partly because other groups could evaluate, reuse, and build on them. The trend also quickly swept the geospatial community: we identify $152$ papers ($2019$--$2025$) in our audited corpus, including papers self-identifying as ``foundation models'' and their self-supervised predecessors. We do not relitigate who may use that title, but instead ask a narrower question a reviewer or downstream user should be able to answer from the published record: which GFM is most performant across diverse or particular tasks by comparable empirical evidence?

Right now, the answer is unclear. For example, we find two papers that report CROMA's mIoU on m-Cashew-Plantation as $30.1$ and $62.2$ (\S\ref{sec:divergence}). A reader deciding whether to use CROMA cannot tell from these papers how differences in decoders and evaluation details affect the comparison. We document 47 such ${\ge}5$-point reported gaps, including 18 ${\ge}10$-point gaps, between papers with the same described model and benchmark grouping in \S\ref{sec:divergence}. This paper addresses this comparability problem and lays out how the community can fix it.

\begin{positionbox}[Position]
\textbf{Nobody knows the current state of the art for geospatial foundation models.} Across the audited corpus, papers do not share a robust evaluation framework. Papers report different numbers for the same model on the same benchmark under the same evaluation protocol. Architectural changes are bundled with pretraining data changes, with no ablations that fix one and vary the other. The GFM literature is missing shared controls during pretraining and evaluation: released weights, a shared set of tests, labels showing which baselines were copied versus rerun, uncertainty reporting, and checks that separate model changes from data changes. The community can and should come together to fix this. Better shared standards will help drive comparability and ultimately innovation that results in better outcomes for end-users in diverse application areas.
\end{positionbox}

\paragraph{Scope.}
The foundation-model title is imported from NLP and computer vision, so it should carry the same minimum standard of evidence that made the title useful there~\citep{bommasani2021opportunities}. We are not claiming that pretrained satellite-imagery backbones are useless; the gap is in how the scientific literature reports and compares them, not in whether the methods work. We do not require every GFM to use public or identical pretraining data; private and diverse data sources are compatible with foundation models when the paper treats data choice as an explicit variable. Our scope is the academic, open-source GFM literature, where public comparability is the main concern. Throughout, ``the field'' refers to GFM literature and its research community, not every remote-sensing or operational geospatial-ML effort.

\paragraph{Contributions.} (1)~We release a 152-paper systematic review with structured per-paper metadata (\S\ref{sec:corpus}). (2)~We describe five troubling trends in GFM papers, following \citet{lipton2019troubling}'s argument that ML papers should make clear what caused an improvement rather than leave readers to guess. (3)~We give six recommendations for authors, reviewers, venues, and benchmark maintainers in \S\ref{sec:rec}, designed to address the troubling trends this paper identifies. We label them \textbf{R1} through \textbf{R6}.

\section{Publication Corpus}
\label{sec:corpus}
To ground our position on GFM comparability in a transparent, reproducible corpus, we construct an audited collection of relevant papers. The supplementary repository contains the paper list, extraction schema, normalized tables, and scripts used for all reported numbers.
We seed this corpus from prior GFM surveys~\citep{lu2024surveygeofm,xiao2024geofmreview} and extend it with two expansion passes: an OpenAlex and Semantic Scholar citation-graph expansion (which adds papers from $2024$--$2025$ that are not covered by the surveys), and a keyword sweep over $2019$--$2023$ remote-sensing self-supervised papers that predate the ``foundation model'' terminology. We add $26$ papers from expert suggestions and new-release monitoring (Appendix~\ref{app:corpus-flow}). Our corpus contains $152$ papers ($2019$--$2025$).

We download the LaTeX source of $140$ of the papers that are available on arXiv and convert the remaining 12 to a structured markdown format from their PDFs using Docling~\citep{livathinos2025docling}. For the LaTeX-source papers, we extract per-paper metadata directly from source using Claude Opus 4.7 and GPT 5.5 Codex; for the PDF-only papers, the Docling markdown is fed to the same extraction pipeline. The extractor writes structured JSON for model, architecture, pretraining method and data, downstream tasks, code and weight release, and key claim. An independent LLM pass flags disagreements, which one author checks against the source papers. Several authors also conduct a five-annotator study of $37$ papers to measure agreement with the extraction and between annotators (Appendix~\ref{app:validation}). The extraction prompt and validation steps are documented in Appendix~\ref{app:validation}; the code and intermediate outputs are included in the supplementary materials. $46\%$ of the $152$ papers explicitly call their proposed model a foundation model in the title, abstract, or contributions. The remaining papers are earlier self-supervised remote-sensing models that prior GFM surveys include alongside more recent foundation-model papers.

We include $2019$--$2025$ papers that introduce or evaluate pretrained Earth-observation representations intended for downstream reuse and either self-identify as foundation models or appear as predecessors in the seed surveys. We exclude paywalled or metadata-poor venues where structured metadata could not be harvested at scale. Figure~\ref{fig:prisma} (Appendix~\ref{app:corpus-flow}) reports exclusion counts and reasons.

Appendix~\ref{app:validation} documents the full extraction and validation pipeline. Appendix~\ref{app:corpus-flow} provides the pipeline for the corpus selection. Appendix~\ref{app:sensitivity} tests restrictions to self-identified foundation models and major venues. All analyses of the corpus in this paper are reproducible from the released code.

\section{Troubling Trends in GFM Comparisons}
\label{sec:findings}

Our analysis identifies five troubling trends in GFM research, each motivating a recommendation for the community. We call these \emph{troubling trends} because repeated reporting choices make model claims harder to understand, echoing the concerns raised by \citet{lipton2019troubling}. Section~\ref{sec:rec} turns these trends into actions for authors, reviewers, and the community.

\subsection{Model weights are not published}
\label{sec:openweights}
Across our publication corpus, $39\%$ of papers release no model weights. Releasing model weights is the minimum precondition for downstream reuse and comparison. Of these, papers making up $19\%$ of the corpus provide a public code repository but no released model artifact, so reuse would require attempting to retrain the model with the authors' codebase (App.~\ref{app:release-audit}). Lack of published model weights is the first troubling trend: before the GFM community can compare models on shared benchmarks or rerun baselines, the model files need to be public.

\begin{recbox}[R1. Release weights under a named license (\S\ref{sec:openweights})]
A pretrained-satellite-imagery model that is meant for reuse should release weights under a named license by camera-ready publication, or explicitly name the constraint that prevents release.
\end{recbox}

\subsection{The field does not have a shared set of core benchmarks}
\label{sec:benchmarks}

\begin{figure}[t]
  \centering
  \begin{subfigure}[t]{0.32\linewidth}
    \includegraphics[width=\linewidth]{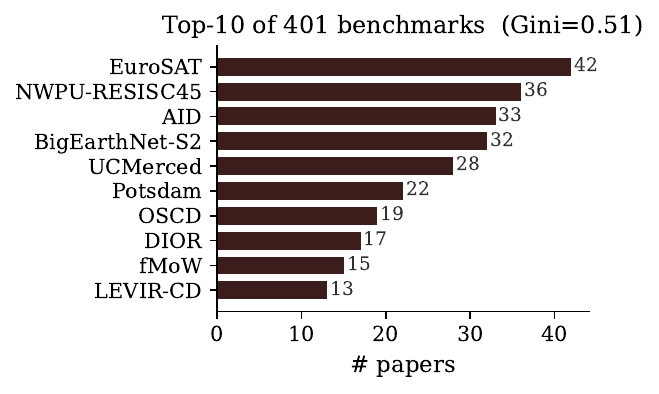}
    \caption{\textbf{Top-10 benchmarks of 401 ranked by paper count.} The top-$3$ (EuroSAT, NWPU-RESISC45, AID) cover only $10.6\%$ of all evaluations; the remaining $89.4\%$ scatters across $398$ benchmarks that mostly appear in one or two papers.}
    \label{fig:bench-left}
  \end{subfigure}\hfill
  \begin{subfigure}[t]{0.32\linewidth}
    \includegraphics[width=\linewidth]{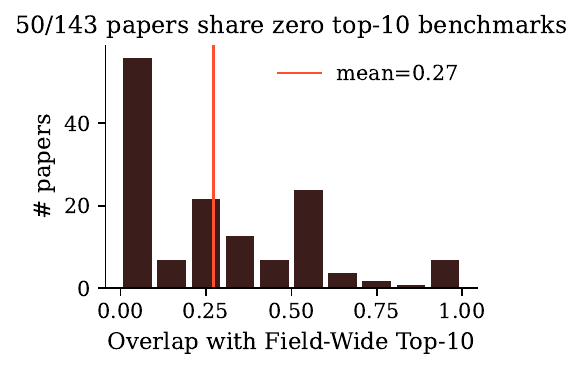}
    \caption{\textbf{Histogram of per-paper overlap with the GFM top-10 benchmarks.} Mean overlap is $0.27$ and $50/143$ papers ($35\%$) sit in the leftmost bin at zero overlap, so more than a third of the corpus shares no evaluation ground with the most-used benchmarks.}
    \label{fig:bench-right}
  \end{subfigure}\hfill
  \begin{subfigure}[t]{0.32\linewidth}
    \includegraphics[width=\linewidth]{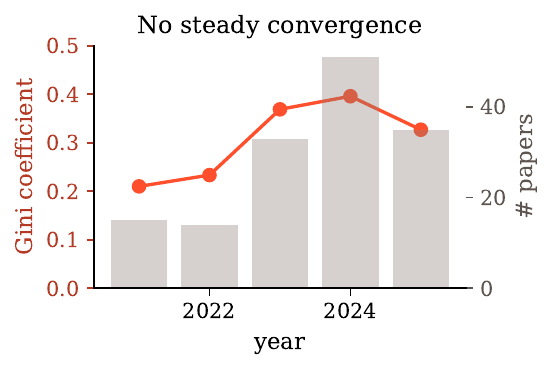}
    \caption{\textbf{Gini coefficient of benchmark usage by publication year} ($0$ means evenly spread, $1$ means a single benchmark dominates). Gini does not steadily increase after $2023$, \emph{so the field is not converging on a shared core over time.}}
    \label{fig:conc-left}
  \end{subfigure}
  \caption{\textbf{How the 152-paper corpus uses benchmarks.} Panel~(a) shows the top-10 benchmarks evaluated in the corpus; panel~(b) shows that $35\%$ of papers do not test on the most-used benchmarks at all; panel~(c) shows that this pattern is not improving over time. Together the three panels say no GFM in the corpus can be ranked literature-wide, because the numbers needed for a fair comparison are not reported on enough shared benchmarks.}
  \label{fig:benchmarks}
\end{figure}

The corpus does not converge on a shared set of benchmarks.
The $152$ papers in our corpus report evaluations on $401$ distinct benchmarks. We determined the number of distinct benchmarks by merging benchmark aliases and excluding evaluations on auxiliary label sources such as the USDA Cropland Data Layer (see full criteria in Appendix~\ref{app:validation}). The corpus has a total of $1{,}046$ evaluation experiments, with an average of $2.6$ evaluations per benchmark.
The three most-used benchmarks (EuroSAT \citep{helber2019eurosat}, NWPU-RESISC45 \citep{cheng2017remote}, AID \citep{xia2017aid}) together account for only $10.6\%$ of all evaluations (Figure~\ref{fig:bench-left}); the remaining $89.4\%$ is spread over $398$ benchmarks, most appearing in only one or two papers. The Gini coefficient---a $[0,1]$ measure of inequality where $0$ indicates evenly distributed usage across benchmarks and $1$ indicates that a single benchmark dominates---is $0.51$ (95\% bootstrap CI $[0.47, 0.55]$), indicating that evaluations are unevenly distributed across benchmarks.

Heavy use of a few benchmarks would be expected as research communities converge on canonical ones. The problem is on the per-paper side. For each paper, we compute the fraction of its downstream benchmarks that overlap the top-10. The mean is $0.27$ (95\% CI $[0.23, 0.32]$), and $50/143$ papers ($35\%$) have \textbf{zero} overlap with the top-10 (Figure~\ref{fig:bench-right}). \emph{Most papers do not evaluate on a shared core, and this is not improving over time.} The year-by-year Gini coefficient is stable after $2023$ (Figure~\ref{fig:conc-left}), and \emph{the count of benchmarks that appear in only one paper grew from $13$ in $2022$ to $98$ in $2025$}.

This indicates no GFM in the corpus can credibly claim a literature-wide ranking from the published record, as the numbers needed to rank them are not reported on enough shared benchmarks for a fair comparison. The GFM literature has its own benchmark-lottery problem, and without the kind of community coordination that has begun to make computer-vision benchmarks more comparable~\citep{koch2021reduced,ott2022saturation}, it is unlikely to fix itself.

\begin{recbox}[R2. Report on a shared core set of evaluation datasets (\S\ref{sec:benchmarks})]
Authors making any pretrained-satellite-imagery comparison should report on shared core evaluations appropriate to their modality and claimed tasks, with a clear protocol, plus additional tests for other aspects of the claimed novelty.
\end{recbox}

\subsection{Reported metric values diverge by tens of points across papers, under supposedly the same evaluation recipe}
\label{sec:divergence}

A field that shares benchmarks should at least agree on reported metric values for the same model-benchmark-protocol tuple. We mine every (model, benchmark, metric, evaluation-protocol, train-regime) tuple in the $152$-paper corpus ($11{,}488$ results after normalization and human-reviewed corrections) and group the results by protocol (fine-tuning, linear probe, kNN probe, zero-shot, few-shot). We then drop generic and classical-ML baselines (random init, generic ImageNet-supervised labels, MLP, from-scratch, LightGBM, XGBoost, SVM, kNN). We also drop detection benchmarks (DOTA \citep{xia2018dota}, DIOR \citep{li2020object}, FAIR1M \citep{sun2022fair1m}) whose shared mean average precision (mAP) metric name conflates DOTA-style oriented-box mAP with COCO-style AP$[.5{:}.95]$ on horizontal boxes. Every remaining disagreement is between papers within a fixed protocol bucket. After these filters, $252$ tuples are reported by ${\ge}2$ papers, and we measure the spread ($\max-\min$) of the reported metric on each.

\begin{figure}[t]
  \centering
  \begin{subfigure}[t]{0.4\linewidth}
    \centering
    \includegraphics[width=\linewidth]{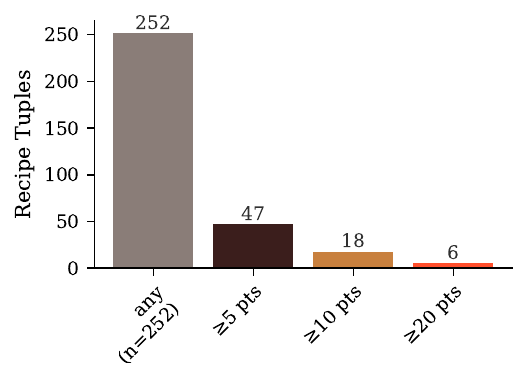}
    \caption{Cross-paper spread at fixed protocol.}
    \label{fig:divergence-left}
  \end{subfigure}\hfill
  \begin{subfigure}[t]{0.58\linewidth}
    \centering
    \includegraphics[width=\linewidth]{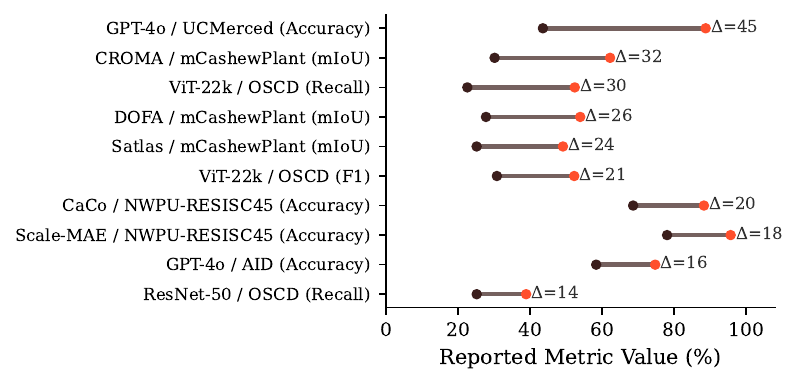}
    \caption{Top-10 reported disagreements.}
    \label{fig:divergence-right}
  \end{subfigure}
  \caption{\textbf{Papers report wildly different numbers for supposedly the same experiment and evaluation recipe.} Recipe tuples are combinations of \texttt{(model, benchmark, metric, eval)}. Across $252$ recipe tuples with matching labels, $47$, $18$, and $6$ have spreads of $\geq\!5$, $\geq\!10$, and $\geq\!20$ points, respectively (\emph{left}); the $10$ largest gaps are shown \emph{right}. For example, \texttt{CROMA} on \texttt{m-Cashew-Plantation} spans $30.1$ to $62.2$ mIoU. Seed variance under a fixed protocol is typically $\sim\!1$ point, so these differences are far larger than what would be expected from run-to-run variation, indicating that \textbf{direct cross-paper comparisons are unreliable}.}
  \label{fig:divergence}
\end{figure}

Of the $252$ multi-paper tuples, $47$ ($18.7\%$) have spread ${\ge}5$~pts, $18$ ($7.1\%$) have spread ${\ge}10$~pts, and $6$ have spread ${\ge}20$~pts (Figure~\ref{fig:divergence}, left). For example, CROMA on m-Cashew-Plantation spans $30.1$ mIoU in DOFA~\citep{xiong2024dofa} to $62.2$ in SatDiFuser~\citep{jia2025can}, a \textbf{32.1-point reported gap}. SatDiFuser uses UPerNet, and the papers do not provide enough detail to establish matching decoders and evaluation details. The largest spread is $45.2$~pts for GPT-4o on UCMerced under zero-shot evaluation. The top-$10$ reported gaps are plotted in Figure~\ref{fig:divergence} (right). These gaps extend beyond isolated outliers. Most multi-paper tuples agree closely (median spread $0.1$~pts), but the \textbf{90th-percentile spread is 8.3~pts}, larger than typical seed variance for classification heads under fixed protocols~\citep{bouthillier2021accountability,bouthillier2019unreproducible}. Establishing whether these gaps reflect replication failures requires matching the evaluation recipes. Variance for segmentation and regression decoders is rarely reported, a common theme for many of these disagreements.

Several plausible failure modes are consistent with the spread. Papers may copy numbers from other papers' tables without annotating that the source used a different train/val/test split, sensor or set of spectral bands, class set, normalization, or adaptation recipe. Papers may rerun baselines with less generous sweeps than the original source and report the rerun as if it were the same protocol. Vision-language papers add prompt templates, verbalizers, API snapshots, and temperature as hidden axes. In every case, the label ``EuroSAT accuracy under linear probe'' specifies less than readers and authors assume. Table~\ref{tab:divtop} (Appendix~\ref{app:divergence}) lists the top-$10$ most divergent tuples and cites the reporting papers, so readers can refer to the source tables directly. \citet{fuller2026badtables}'s ``BAD TABLES'' talk catalogs the same confounding variables at the architecture level: patch size, image size, channel groupings, and pretraining schedule each shift downstream accuracy by tens of points across experiments with the same model.

\begin{recbox}[R3. Annotate every copied baseline number (\S\ref{sec:divergence})]
Every result-table entry should be marked as either \rerunmark~\textsc{rerun}, with the configuration disclosed, or \copiedmark~\textsc{copied}, with the source paper and source protocol cited.
\end{recbox}

\begin{recbox}[R4. Report seed variability on headline comparisons (\S\ref{sec:divergence})]
Headline comparisons should show how stable the result is when that is practical, using repeated-run mean$\pm$std for affordable evaluations, or a clear note that the result comes from one run when repeats are too expensive. This is especially important for benchmarks where results are known to vary greatly across random seeds.
\end{recbox}

\subsection{Aggregated benchmarks provide dataset bundles, but not evaluation harnesses}
\label{sec:harness}

The LLM community converged on \emph{evaluation harnesses}, not just collections of benchmark datasets. For example, \texttt{lm-evaluation-harness}~\citep{gao2024lmeval} is the canonical tool that powers the Open LLM Leaderboard~\citep{fourrier2024openllm}: a single Python package every model owner runs, with versioned task definitions (e.g., MMLU \texttt{v0.0} vs.\ \texttt{v0.1}), reference protocol implementations, automated submission checks through continuous integration, and a common task-config format that lets any third party reproduce a reported number from a model name and a task tag. HELM~\citep{liang2023helm} provides multi-metric evaluation across $87$ scenarios under a continuously hosted leaderboard. BIG-bench~\citep{srivastava2023bigbench} provides $200{+}$ tasks under a unified API. Computer vision also has common harnesses: outside of ImageNet~\citep{deng2009imagenet}, VTAB~\citep{zhai2019vtab} defines a $19$-task transfer-learning protocol with reference implementations.

The geospatial domain lacks this infrastructure. What the GFM community has are \emph{dataset bundles}: GitHub repositories with curated task splits, reference dataloaders, and example training scripts. GEO-Bench~\citep{lacoste2023geobench,simumba2025geo} provides fixed splits and a public toolkit. PANGAEA~\citep{marsocci2024pangaea} provides a unified codebase that runs encoders across a fixed task list. FoMo-Bench~\citep{bountos2024fomobench} curates a forest-monitoring task list. PhilEO Bench~\citep{fibaek2024phileobench} is a paper proposing a list of datasets with no harness. TorchGeo~\citep{stewart2025torchgeo} has broad dataset and transform coverage for geospatial ML, but it is a general-purpose library rather than a probing harness with canonical GFM submissions and leaderboards. None of these are evaluation harnesses like those available for LLMs or CV. Each is a self-contained repository or toolkit where a researcher forks and writes their own training loop on curated splits; cross-paper protocols are not versioned, submissions are not tested with CI, and there is no canonical tool the whole community runs.

We argue that disagreements in \S\ref{sec:divergence} occur even on shared benchmarks because there is no common evaluation harness. Even when two GFM papers evaluate on the same dataset from the same bundle, the results remain incomparable as long as evaluation protocol details such as optimizer choice, head learning rate, eval crop, and Jaccard-averaging scheme (macro vs.\ weighted-per-class IoU) are inconsistent. \emph{Curating more datasets does not close that protocol gap.} What the GFM literature is missing is a third-party evaluation harness that is a versioned, openly maintained tool that every model owner runs to produce a reported number, and that any reviewer can rerun end-to-end from a model identifier.

The shared harness should evaluate models on datasets that reflect their intended uses. Existing remote-sensing datasets often reflect where labels were convenient to collect, not the full distribution of operational tasks, geographies, sensors, and labeling practices. Patch-level or image-level scores also may not predict map-level accuracy or user-facing utility. Spatial autocorrelation between training and validation samples can inflate remote-sensing accuracy estimates~\citep{karasiak2022spatial,kattenborn2022spatially}, so an evaluation that is reproducible can still be the wrong evaluation for a deployment. The right target is therefore not any one specific benchmark (EuroSAT is just a common example); it is a shared core for sanity-checking model comparisons, plus Earth-observation-native extension axes for sensor, modality, temporal, geographic, label-quality, and map-level claims.

\begin{recbox}[R5. Create a shared third-party evaluation harness (\S\ref{sec:harness})]
Community benchmark maintainers and venues should build one shared evaluation tool that every model owner can run, with fixed task definitions and automatic submission checks.
\end{recbox}

\subsection{Architecture and pretraining-data improvements are confounded}
\label{sec:pretrain}

\begin{wrapfigure}{r}{0.45\linewidth}
  \vspace{-6pt}
  \centering
  \includegraphics[width=\linewidth]{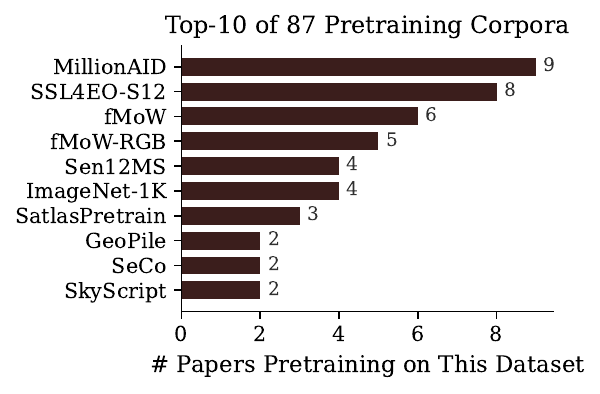}
  \caption{{\textbf{Top-10 (of 87) named pretraining datasets} across the 126 corpus papers that name one. \texttt{MillionAID} leads at just 9 papers ($\sim$5.9\% of 152); \texttt{SSL4EO-S12} (8), \texttt{fMoW} (6), and \texttt{fMoW-RGB} (5) follow.
  }}
  \label{fig:pretrain}
\end{wrapfigure}

When a paper changes both the model and the pretraining data, readers cannot tell which change caused the gain unless one is held fixed. Attributing gains requires controls, not identical pretraining data. A foundation model does not need to share pretraining data---private, proprietary, and diverse pretraining data are acceptable when a novel pretraining dataset is the claimed contribution or when a paper does not ask readers to attribute gains to architecture alone. The problem in the GFM corpus is that methodological claims (architectural changes, new self-supervised objectives) are often impossible to separate from pretraining-data changes without an ablation that fixes one and varies the other. \citet{corley2024revisiting} show that apparent GFM gains over supervised ImageNet baselines on common remote-sensing classification benchmarks shrink or vanish once the pretraining and downstream distributions and image sizes are held constant. This provides evidence that the architecture-vs-pretraining-data confound can hide effects the corpus would benefit from disentangling. \citet{kaur2026pretrainwhere} show that different pretraining datasets can shift downstream accuracy by margins comparable to or larger than gains attributed to architectural novelty.

To compare pretraining data across papers, we merge pretraining dataset aliases and drop unnamed or misnamed pretraining datasets (see full filter list in Appendix~\ref{app:validation}). We count $87$ distinct primary pretraining corpora across $126$ papers that name a specific dataset (the remaining $26$ papers describe their pretraining data only generically, e.g., Sentinel-2). Some papers pretrain on a single canonical dataset, but many build a mixture from multiple sources and give it a single name. For example, RS5M \citep{zhang2024rs5m} is built from LAION \citep{schuhmann2022laion} + CC3M \citep{sharma2018conceptual} + CC12M \citep{changpinyo2021conceptual} + others. AnySat's GeoPlex \citep{astruc2025anysat} wraps TreeSatAI-TS \citep{astruc2024omnisat} + FLAIR \citep{garioud2023flair} + PLANTED \citep{pazos2024planted} + PASTIS-HD \citep{astruc2024omnisat}. GeoPile \citep{mendieta2023towards} wraps MillionAID \citep{Long2021DiRS} + SEN12MS \citep{schmitt2019sen12ms} + MDAS \citep{hu2023mdas}. In these cases, we only increment the count of the individual datasets, not the wrapper name. Even so, the most-used primary pretraining dataset, MillionAID \citep{Long2021DiRS}, appears in only $9$ papers, followed by SSL4EO-S12 \citep{wang2023ssl4eo} ($8$), fMoW \citep{christie2018functional} ($6$), and fMoW-RGB ($5$) (Figure~\ref{fig:pretrain}). The actual scene composition behind named sensor labels (Sentinel-1/2, Landsat, NAIP) is split across dozens of overlapping derived pretraining datasets (SSL4EO-S12 \citep{wang2023ssl4eo}, fMoW-Sentinel \citep{christie2018functional}, MMEarth \citep{nedungadi2024mmearth}, MajorTOM-Core \citep{francis2024major}, SatlasPretrain \citep{bastani2023satlaspretrain}) whose intersection cannot be audited from the papers alone.

Counting individual datasets also misses differences in how papers combine them. Two papers that name the same pretraining dataset may still pretrain on different data: a paper that pretrains on BigEarthNet alone is not equivalent to one that pretrains on a custom mixture in which BigEarthNet is one source among many. Both list BigEarthNet, but the resulting models see different data. We compute the \emph{full} pretraining set for each paper (the deduplicated set of all named datasets the paper pretrains on), and compute how often two papers' full sets are identical. Of $126$ papers with an extractable pretraining dataset, only $32$ ($25\%$) share their full configuration with at least one other paper. The other $94$ papers each pretrain on a set of data no other work uses. The most common shared pretraining set is MillionAID alone, used by $7$ papers, followed by SSL4EO-S12, used by $6$ papers. Outside that handful of clusters, no two papers in the corpus have run the same pretraining recipe.

\begin{recbox}[R6. Disentangle pretraining dataset from architecture and algorithm (\S\ref{sec:pretrain})]
Any new GFM that changes more than one of \textit{a) pretraining data, b) architecture (including subtle choices like patch size), and c) algorithm} should include controlled comparisons that isolate the claimed change. Without these controls, weak results can hide behind stronger changes, making it unclear which modification actually produced the improvement.
\end{recbox}

\section{Recommendations}
\label{sec:rec}

The recommendation boxes in \S\ref{sec:findings} propose actions to address each trend: release reusable weights, evaluate on a shared core, distinguish copied from rerun baselines, report uncertainty, build a shared evaluation tool, and separate model changes from data changes. Most of these recommendations are actions that authors can (and, we argue, should) start doing today. Our contribution is the GFM-specific evidence and adaptations of existing evaluation practices (Appendix~\ref{app:precedents}).

\begin{description}[leftmargin=0em, style=sameline]
    \item \textbf{R1: Release reusable weights or name the constraint.} A model intended for reuse should provide weights under a named, permissive-by-default license by camera-ready publication. $39\%$ of the corpus currently releases no weights, which blocks reuse and the cross-paper checks needed for R3 and R6. When sensor licensing, rules about where data can be stored, restrictions on sharing technology across borders, or partner restrictions prevent release, authors should name the reason rather than leave readers guessing.

    \item \textbf{R2: Evaluate on a minimum set of shared core benchmarks.} Authors should use one versioned harness with optical, SAR, multimodal/vision-language, time-series, and dense map-level tracks, reporting on the tracks their model claims to handle. A shared core across tracks applies only where model inputs and claimed tasks permit. Initially, authors could prioritize the most common benchmarks in our corpus (see Figure~\ref{fig:bench-left}). Alternatively (or in addition), authors could prioritize the GEO-Bench and PANGAEA benchmark collections. We acknowledge, though, that older or more popular benchmarks are not automatically the right ones. The GFM research and end-user community should convene to identify the core benchmarks to prioritize in evaluations. Model producers, benchmark maintainers, and operational users should jointly select and maintain the core benchmarks.

    \item \textbf{R3: Mark copied versus rerun baselines.} Every baseline entry should be marked as \copiedmark~\textsc{copied} or \rerunmark~\textsc{rerun}. Copied rows need the source paper and source protocol; rerun rows need the new configuration. Tables in which the proposed method is rerun but all baselines are copied without protocol notes are not fair comparisons. If a rerun is far from the source number, the paper should flag that rather than quietly replace the old result.

    \item \textbf{R4: Report uncertainty where it is affordable.}
    Whenever possible (e.g., for cheap heads and headline classification comparisons), authors should report mean$\pm$std over at least three complete repeats, varying initialization, data order, and augmentation together~\citep{bouthillier2021accountability}. Separate ablations of each source of randomness are not required. For expensive segmentation, regression, time-series, or map-level runs, a narrower first step is enough: say whether the headline result is from one run, state the hyperparameters and main sources of randomness, and avoid claiming a clear improvement unless the gain is likely bigger than normal run-to-run differences. Computer-vision and ML reproducibility work has shown that small implementation and seed choices can change reported gains~\citep{bouthillier2019unreproducible}, and the GFM literature has not yet measured its own run-to-run variation well enough to ignore this concern.

    \item \textbf{R5: Create a shared third-party evaluation harness.}
    The missing community layer is a third-party-maintained harness with versioned protocols, a common submission format, and automated checks that any model owner runs to produce a reported number (\S\ref{sec:harness}). GEO-Bench~\citep{lacoste2023geobench,simumba2025geo}, PANGAEA~\citep{marsocci2024pangaea}, and related dataset bundles are the starting point; TerraTorch~\citep{gomes2025terratorch} and PANGAEA take steps in this direction, but the GFM literature still lacks one shared tool tested with CI that runs the same protocol for everyone.

    \item \textbf{R6: Separate model changes from pretraining-data changes.} If a paper introduces both a new pretraining dataset and a new architecture or self-supervised objective, it should include at least one comparison that fixes the data source to a public choice (e.g., SSL4EO-S12, MillionAID, fMoW, or MajorTOM-Core) at a similar token or image budget. At minimum, papers using the same sensor should compare within that sensor family. Variation in pretraining data is welcome; it should be the thing being studied, but not a hidden reason for a gain~\citep{kaur2026pretrainwhere}. The ablation does not need to be a frontier-scale run. Re-pretraining a ViT-L on a billion-image dataset is beyond many labs, and R6 should not be read as ``every paper must pretrain on the same handful of datasets.'' If the paper claims an architectural or objective improvement, include at least one controlled run on a canonical public pretraining dataset or a documented, deduplicated, region-balanced subset.

\end{description}

\paragraph{For conferences, workshops, program committees, and the community at large.} Machine learning venues (e.g., ICML, ICLR, NeurIPS, CVPR, ICCV, ECCV) and geospatial-specific venues (e.g., EarthVision at CVPR, IGARSS, and ISPRS) should treat R1--R6 as default expectations and non-compliance as a substantive review concern. Reviewers cannot verify artifacts that are only promised for camera-ready, so venues should require anonymous weights, code, or executable evaluation artifacts at submission when those artifacts support the paper's central claims; otherwise, the paper should state that the numbers cannot be independently checked during review. A reviewer need not require all six recommendations to be met. The key question is which missing check confounds the paper's claim. We encourage reviewers to use the following checklist to help identify issues rather than treating it as a pass/fail requirement.

\begin{checklistbox}[Reviewer's checklist (mapped to R1--R6)]
\begin{itemize}[label=\checkedbox,leftmargin=*,itemsep=2pt,topsep=2pt]
  \item \textbf{C1 [R1], released model files:} Are weights shared under a named license by camera-ready, or does the paper clearly state why they cannot be released?
  \item \textbf{C2 [R2], shared evaluation:} Are there at least three benchmarks from a shared core set (for example, the $10$ most-evaluated benchmarks across the corpus, or a common bundle like GEO-Bench or PANGAEA) and is the protocol stated explicitly? %
  \item \textbf{C3 [R3], result source:} Is every baseline row marked \rerunmark~\textsc{rerun} with the configuration disclosed, or \copiedmark~\textsc{copied} with the source paper and its protocol cited? For headline benchmarks, does the paper match a published prior number under an explicit protocol or rerun end-to-end?
  \item \textbf{C4 [R4], uncertainty reporting:} Does the headline table report mean$\pm$std over at least three complete repeats where feasible, or clearly state that the result is single-run?
  \item \textbf{C5 [R5], shared harness:} Does the paper name the evaluation harness version and release its configuration, or explain why the harness does not support the evaluation?
  \item \textbf{C6 [R6], pretraining-data control:} If the paper introduces both a new architecture or self-supervised objective \emph{and} a new pretraining dataset, is there a comparison fixing the pretraining dataset to a shared public choice?
\end{itemize}
\end{checklistbox}

\section{Alternative Views}
\label{sec:alt}

\paragraph{``The field is young; concentration and divergence will self-correct.''} Current GFM surveys~\citep{lu2024surveygeofm,xiao2024geofmreview} seem to assume that comparisons will improve as the field matures, while the GEO-Bench authors~\citep{lacoste2023geobench,simumba2025geo} propose a shared leaderboard to help make that happen. We agree that the leaderboard route is the right one. However, the failures we measure ($32.1$-pt CROMA disagreement, $35\%$ zero top-10 overlap, $75\%$ of papers pretraining on a unique configuration) will not fix themselves with time alone. A new model can meet R1--R6 at release, and Figure~\ref{fig:conc-left} shows benchmark concentration has not improved since $2023$. The risk of waiting is that GFM papers keep reporting progress without knowing what drove it: the model design, the data, the protocol, run-to-run noise, or the benchmark choice. A shared standards document gives authors and reviewers a common reference before informal habits become hard to change.

\paragraph{``Geospatial data and tasks are heterogeneous; do not shoehorn remote sensing into a CV or LLM box.''}
\citet{rolf2024position,marsocci2024pangaea,fuller2026badtables} argue that input heterogeneity (multispectral, SAR, multi-resolution, multi-temporal) should set remote-sensing-native standards. We are not arguing for one mandated input format. Instead we argue that input-axis heterogeneity itself does not explain disagreement within an axis. The CROMA/m-Cashew-Plantation disagreement (\S\ref{sec:divergence}) occurs within a single benchmark and described evaluation recipe. We ask for controls plus diversity: a small shared core, with explicit extension axes for sensor-, modality-, and time-series-specific tasks. %

\paragraph{``A shared core encourages benchmark overfitting.''} A fixed core can narrow research priorities and reward benchmark-specific gains~\citep{dehghani2021benchmarks}. It should be versioned and supplemented with geographic and temporal holdouts, while allowing new tasks. Repeated-run costs and legitimate commercial or security restrictions also limit what R1--R6 can require.

\paragraph{``These problems are driven by bad actors, not the whole field.''}
Our corpus already removes some noise by excluding metadata-poor venues and obvious non-reuse papers, so the trends are not driven by an anything-goes scrape. Still, we acknowledge that a stricter ``top papers only'' check could show different trends, but there is no obvious filter for this. Restricting the corpus further to major venues reduces the share without released weights, but all other troubling trends remain (Appendix~\ref{app:sensitivity}). Our claim is about the public record a reviewer, leaderboard maintainer, or downstream user actually sees. \emph{Those readers should not need informal field knowledge to know which papers are careful, which baselines were copied, or which protocol was rerun.} If the reliable signal exists only inside a small social circle, it is not yet serving as public evidence.

\paragraph{``Model weights or pretraining data sometimes can't be open-sourced because of private or proprietary data constraints.''}
We agree that proprietary models that are closed-weight or closed-data can still add value to the research community; AlphaEarth~\citep{brown2025alphaearth} is one such example in geospatial foundation models. R1 recommends that authors state such constraints if they are blocking the release of model weights or pretraining data, but if such constraints are not present, all artifacts should be released.
We do, however, contend that open-source practices are largely beneficial for the goals stated in this paper, in line with Donoho's notion of ``frictionless reproducibility''~\citep{donoho2024data}.

\section{Conclusion}
\label{sec:conclusion}

We argue that nobody knows the current state of the art for geospatial foundation models because the published literature does not yet share the controls and standards a reader needs to compare them. The evidence from our $152$-paper audit supports this claim: papers do not share enough benchmarks, the same model on the same benchmark gets very different reported scores across papers, and pretraining setups vary so much that model gains are hard to attribute without controls. We suggest six recommendations for authors and the research community to correct the troubling trends identified in this paper, focusing on actions authors can take now or in the near term. We also suggest a reviewer checklist to help identify when the experiments or contributions in a GFM paper are cause for concern.

\paragraph{Limitations.} Our analysis relies on LLM-based automated extraction, public APIs, and public data. Docling converted PDF-only papers to markdown, and Claude Opus 4.7 and GPT 5.5 Codex extracted structured metadata from all sources, followed by one author checking fields where two LLM passes disagreed and spot checks by several authors. A separate five-annotator study covers $37$ papers (Appendix~\ref{app:validation}). Details of the extraction are described in App.~\ref{app:validation}.

\paragraph{What the future looks like if we are successful.} A user who needs a GFM for crop mapping, flood response, or building detection can open a maintained leaderboard, choose the task, sensor, region, and date range they care about, and trust that the top entries were run under the same recipe rather than stitched together from incomparable paper results. A researcher releasing a new model does not need to compare against every backbone from the last seven years just because nobody knows which few baselines still matter; they run a shared community-maintained harness, mark any extra copied numbers as \copiedmark~\textsc{copied}, and spend the rest of the paper explaining what actually changed. A reviewer can ask whether a missing check changes the claim, not whether the whole table is real. A leaderboard maintainer can rerun a result from a versioned model config, see when a protocol changed, and flag scores whose uncertainty or data controls are missing. In that future, GFMs can still be diverse in architecture, scale, modality, and application, but their public evidence becomes boring in the best way: comparable enough that the interesting arguments are about models and data, not table archaeology. \emph{\textbf{To know which geospatial foundation model is best, we must first make them comparable.}}

\section*{Acknowledgments}

We thank Konstantin Klemmer for his careful and in-depth review of the manuscript. His detailed comments substantially improved the clarity and presentation of this work.

\section*{Funding}

No specific funding was received for this collaborative work.

\bibliographystyle{plainnat}
\bibliography{references}

\clearpage
\appendix

\section{Reproducibility}
\label{app:repro}

The supplementary repository contains the $152$-paper corpus, extraction prompts, normalization code, harvested results, and figure code; a top-level \texttt{Makefile} regenerates everything end-to-end.

\section{Extraction Pipeline}
\label{app:validation}

We extract structured records directly from the LaTeX sources of the arXiv papers using Claude Opus 4.7 and GPT 5.5 Codex, and we run the same pipeline on the remaining PDF-only papers after converting them to structured markdown with Docling~\citep{livathinos2025docling}. Two independent LLM passes extract the fields we analyze. One author checks every flagged disagreement against the source paper, and several authors spot-check additional papers. We normalize names to deduplicate repeated models, benchmarks, and pretraining datasets. We also remove labels that are not held-out evaluation sets and generic dataset descriptions that do not name a specific pretraining corpus. The normalization code and validator are provided with the released code. We also conducted a separate five-annotator study on $37$ papers, described below.

\paragraph{Annotator study.} Five annotators independently checked benchmarks, pretraining corpus, code release, weight release, and one sampled result for each of $37$ randomly selected papers and returned $921$ field judgments. Agreement with the extraction was $822/921=89.3\%$, while mean pairwise agreement between annotators was $85.4\%$. One author adjudicated disagreements after all responses were collected. Result disagreements predominantly concerned protocol and training-fraction labels rather than incorrect metric values after source verification.

\begin{table}[ht]
  \centering
  \small
  \caption{\textbf{Annotators usually endorsed the extraction.} Pairwise agreement compares annotators' four verdicts (agree, partial, disagree, unsure). Majority endorsement excludes ties.}
  \label{tab:annotators}
  \begin{tabular}{@{}lcc@{}}
    \toprule
    \textbf{Field} & \textbf{Pairwise Agreement} & \textbf{Majority Endorsement} \\
    \midrule
    Benchmarks & $96.2\%$ & $37/37$ ($100\%$) \\
    Pretraining Corpus & $85.7\%$ & $37/37$ ($100\%$) \\
    Code Release & $88.6\%$ & $34/37$ ($91.9\%$) \\
    Weight Release & $87.6\%$ & $34/36$ ($94.4\%$) \\
    Sampled Result & $68.1\%$ & $30/35$ ($85.7\%$) \\
    \bottomrule
  \end{tabular}
\end{table}

\section{Weight \& Code Release Audit}
\label{app:release-audit}

Each paper's weight-release flag is set by extracting the explicit claim in the PDF or LaTeX source and, when that is unclear, checking whether the release is hosted publicly elsewhere. Of the $152$ papers, $93$ release weights, $26$ explicitly do not, and weights for the remaining $33$ could not be found. In total, $59/152 = 39\%$ release no model weights. Of these $59$, $29$ ($19\%$ of the corpus) provide a public code repository with no released model artifact.

\begin{table}[t!]
  \centering
  \scriptsize
  \setlength{\tabcolsep}{3pt}
  \renewcommand{\arraystretch}{1.13}
  \renewcommand{\tabularxcolumn}[1]{m{#1}}

  \caption{\textbf{Top-10 cases where the same evaluation gives different scores.}
  Each row groups reports for the same model, benchmark, metric, and stated protocol.
  \emph{Min} and \emph{Max} are the lowest and highest reported scores. \# Papers is the
  number of papers reporting that same described evaluation. The cited papers report the minimum and maximum, respectively. The last column provides likely missing setup details.}
  \label{tab:divtop}

  \begin{tabularx}{\linewidth}{@{}
    >{\raggedright\arraybackslash}m{1.25cm}
    >{\centering\arraybackslash}m{2.15cm}
    >{\centering\arraybackslash}m{1.10cm}
    >{\centering\arraybackslash}m{0.55cm}
    >{\centering\arraybackslash}m{0.55cm}
    >{\centering\arraybackslash}m{0.55cm}
    >{\centering\arraybackslash}m{2.0cm}
    >{\centering\arraybackslash}X
  @{}}
    \toprule
    \textbf{Model} &
    \textbf{Benchmark} &
    \textbf{Metric / Eval} &
    \textbf{\shortstack{\#\\Papers}} &
    \textbf{Min} &
    \textbf{Max} &
    \textbf{Papers} &
    \textbf{Likely Missing Setup Detail} \\
    \midrule

    GPT-4o &
    UCMerced &
    Acc. / 0-Shot &
    2 &
    43.5 &
    88.8 &
    RingMo-Agent~\cite{hu2025ringmo}; EarthDial~\cite{soni2025earthdial} &
    Identical prompts, API snapshot, scoring, and samples are not established. \\

    \midrule

    CROMA \cite{fuller2023croma} &
    m-Cashew &
    mIoU / Linear &
    4 &
    30.1 &
    62.2 &
    DOFA~\cite{xiong2024dofa}; SatDiFuser~\cite{jia2025can} &
    Decoder details are weakly described. \\

    \midrule

    ViT-22k &
    OSCD &
    Recall / FT &
    2 &
    22.5 &
    52.4 &
    GFM~\cite{mendieta2023towards}; SpectralGPT~\cite{Hong2023SpectralGPTSRA} &
    GFM describes $192\times192$-pixel tiles, whereas SpectralGPT describes $128\times128$-pixel crops with 50\% overlap. It is unclear why this difference would produce such a large performance gap. \\

    \midrule

    DOFA~\cite{xiong2024dofa} &
    m-Cashew &
    mIoU / Linear &
    3 &
    27.7 &
    53.9 &
    TerraFM~\cite{danish2025terrafm}; SatDiFuser~\cite{jia2025can} &
    Decoder details are weakly described. \\

    \midrule

    Satlas~\cite{bastani2023satlaspretrain} &
    m-Cashew &
    mIoU / Linear &
    2 &
    25.1 &
    49.1 &
    TerraFM~\cite{danish2025terrafm}; SatDiFuser~\cite{jia2025can} &
    TerraFM reports the same Satlas score as the Galileo paper~\cite{tseng2025galileo}, but it is unclear which decoder produced that score. \\

    \midrule

    ViT-22k &
    OSCD &
    F1 / FT &
    3 &
    30.7 &
    52.2 &
    GFM~\cite{mendieta2023towards}; SpectralGPT~\cite{Hong2023SpectralGPTSRA} &
    The papers do not establish that they use the same evaluation recipe. \\

    \midrule

    CaCo &
    NWPU-RESISC45 &
    Acc. / FT &
    2 &
    68.6 &
    88.3 &
    CtxMIM~\cite{zhang2025ctxmim}; SelectiveMAE~\cite{wang2025harnessing} &
    CtxMIM's head-fine-tuning description is ambiguous. \\

    \midrule

    Scale-MAE~\cite{reed2023scale} &
    NWPU-RESISC45 &
    Acc. / FT &
    7 &
    78.0 &
    95.7 &
    CtxMIM~\cite{zhang2025ctxmim}; Scale-MAE~\cite{reed2023scale} &
    Freezing is unclear in CtxMIM. The same score appearing in several papers does not show that each reran the experiment. \\

    \midrule

    GPT-4o &
    AID &
    Acc. / 0-Shot &
    2 &
    58.3 &
    74.7 &
    RingMo-Agent~\cite{hu2025ringmo}; EarthDial~\cite{soni2025earthdial} &
    Identical samples, prompts, answer mapping, and API snapshot are not established. \\

    \midrule

    ResNet-50 &
    OSCD &
    Recall / FT &
    3 &
    25.1 &
    38.9 &
    GFM~\cite{mendieta2023towards}; SpectralGPT~\cite{Hong2023SpectralGPTSRA} &
    GFM labels the model ResNet-50, but its scores match a ResNet-18 experiment in SeCo. Which model produced them is unclear. \\

    \bottomrule
  \end{tabularx}
\end{table}

\section{Reported-Number Divergence: Harvest Details}
\label{app:divergence}

Table~\ref{tab:divtop} groups reported numbers by the strictest combination we can extract from each source paper: same model, same benchmark, same metric, same evaluation regime (linear / kNN / fine-tune / zero-shot / few-shot), and same training fraction. Anything outside that combination is unpinned, including the linear-probe recipe, the fine-tune recipe used to rerun baselines, the GEO-Bench probing pipeline used to evaluate a third-party backbone, and the prompt / API version / temperature for vision-language models. The source papers do not consistently establish that those choices match. The remaining filters (we keep only full-train results, drop classical-ML baselines, and exclude detection benchmarks where ``mAP'' conflates different definitions, e.g., mAP, mAP@50, and oriented-object-detection mAP) are in the released code.

\begingroup
\captionsetup{skip=3pt}
\section{Corpus Selection}
\label{app:corpus-flow}

Figure~\ref{fig:prisma} summarizes corpus selection from the archived discovery lists. We deduplicate by arXiv ID, DOI, and normalized title, then screen for the inclusion rule in \S\ref{sec:corpus}. Searches used the terms ``geospatial foundation models'', ``remote sensing'', ``earth observation'', ``self-supervised learning'', ``contrastive learning'', ``masked image modeling'', and ``pretraining''. The API snapshots date to May 7, 2026.

\par\noindent\begin{minipage}{\linewidth}
  \centering
  \begin{tikzpicture}[
    box/.style={draw=posblue!60, fill=posbg, rounded corners=2pt, line width=0.5pt, text width=5.7cm, align=center, inner sep=3pt, font=\footnotesize},
    flow/.style={->, draw=posblue, line width=0.55pt}
  ]
    \node[box] (identified) at (0,0) {\textbf{Records Identified: 296}\\Seed Lists: 135\\OpenAlex Citation / Snowball: 41 / 24\\Semantic Scholar: 44\\Keyword Sweep: 52};
    \node[box] (duplicates) at (6.8,0) {Duplicate Records Removed: 47};
    \node[box] (unique) at (0,-1.3) {\textbf{Unique Papers: 249}};
    \node[box] (unreviewed) at (6.8,-1.3) {Not Retrieved: 7};
    \node[box] (assessed) at (0,-1.95) {\textbf{Assessed for Eligibility: 242}};
    \node[box] (excluded) at (6.8,-1.95) {Excluded With Reasons: 116};
    \node[box] (included) at (0,-2.6) {\textbf{Included From Recorded Searches: 126}};
    \node[box] (other) at (6.8,-2.6) {Expert / New-Release Monitoring: 26};
    \node[box, fill=recbg, draw=recred] (final) at (0,-3.3) {\textbf{Final Corpus: 152 Papers}};
    \draw[flow] (identified) -- (duplicates);
    \draw[flow] (identified) -- (unique);
    \draw[flow] (unique) -- (unreviewed);
    \draw[flow] (unique) -- (assessed);
    \draw[flow] (assessed) -- (excluded);
    \draw[flow] (assessed) -- (included);
    \draw[flow] (included) -- (final);
    \draw[flow] (other.south) |- (final.east);
  \end{tikzpicture}
  \captionof{figure}{\textbf{PRISMA-style reconstruction of the 152-paper corpus.} Exclusions comprise $38$ papers screened as not foundation models, $19$ papers outside the time window, $18$ downstream applications, $13$ general CV methods, $12$ benchmark/dataset releases, $10$ surveys, $4$ out-of-domain papers, and $2$ weather/climate papers.}
  \label{fig:prisma}
\end{minipage}\par
\newpage
\section{Corpus Sensitivity}
\label{app:sensitivity}

The self-identified subset matches foundation-model terminology in titles, model names, extracted contributions, or Semantic Scholar abstracts/summaries. The major-venue subset includes NeurIPS, ICML, ICLR, CVPR, ICCV, ECCV, AAAI, TPAMI, TGRS, ISPRS J-PRS, and JSTARS, excluding workshops. Venues come from Semantic Scholar, with paper metadata as a fallback.

\par\noindent\begin{minipage}{\linewidth}
  \centering
  \small
  \setlength{\tabcolsep}{4pt}
  \captionof{table}{\textbf{The troubling trends persist under narrower corpus definitions.} Overlap uses the full-corpus top-$10$ benchmarks throughout. Denominators include papers with the required metadata or recipe tuples reported by at least two papers within each subset. Spreads are in percentage points.}
  \label{tab:corpus-sensitivity}
  \begin{tabularx}{\linewidth}{@{}Xrrr@{}}
    \toprule
    \textbf{Statistic} & \textbf{Full (152)} & \textbf{Self-Identified (70)} & \textbf{Major Venue (87)} \\
    \midrule
    No Weights Released & $59/152$ ($38.8\%$) & $26/70$ ($37.1\%$) & $26/87$ ($29.9\%$) \\
    Benchmarks / Evaluations & $401/1{,}046$ & $299/620$ & $315/672$ \\
    Zero Top-10 Overlap & $50/143$ ($35.0\%$) & $27/66$ ($40.9\%$) & $25/86$ ($29.1\%$) \\
    Spread $\geq 5$ Points & $47/252$ ($18.7\%$) & $34/163$ ($20.9\%$) & $21/106$ ($19.8\%$) \\
    Spread $\geq 10$ Points & $18/252$ ($7.1\%$) & $12/163$ ($7.4\%$) & $9/106$ ($8.5\%$) \\
    90th-Percentile Spread & $8.3$ & $8.3$ & $8.2$ \\
    Unique Pretraining Configuration & $94/126$ ($74.6\%$) & $47/58$ ($81.0\%$) & $62/75$ ($82.7\%$) \\
    \bottomrule
  \end{tabularx}
\end{minipage}\par
\section{NLP and Computer-Vision Precedents}
\label{app:precedents}

\par\noindent\begin{minipage}{\linewidth}
  \centering
  \small
  \renewcommand{\arraystretch}{1.0}
  \captionof{table}{\textbf{R1--R6 adapt established evaluation practices to Earth observation.} The adaptations account for sensor differences, transfer across geography and time, and practical limits on release and computation.}
  \label{tab:precedents}
  \begin{tabularx}{\linewidth}{@{}>{\raggedright\arraybackslash}p{0.20\linewidth}>{\raggedright\arraybackslash}p{0.31\linewidth}>{\raggedright\arraybackslash}X@{}}
    \toprule
    \textbf{Recommendation} & \textbf{NLP/CV Precedent} & \textbf{GFM Adaptation and Limits} \\
    \midrule
    R1: Release Weights & Reusable BERT and CLIP checkpoints~\citep{devlin2019bert,radford2021learning}. & Name sensor-license, commercial, or security restrictions when weights cannot be released. \\
    \midrule
    R2: Shared Core & VTAB's task suite and HELM's multi-scenario evaluation~\citep{zhai2019vtab,liang2023helm}. & Use modality-specific tracks and geographic/temporal tests. A fixed core can narrow research priorities. \\
    \midrule
    R3: Baseline Sources & Shared reference evaluations in \href{https://github.com/huggingface/pytorch-image-models}{timm} and lm-evaluation-harness~\citep{gao2024lmeval}. & Mark copied versus rerun scores and state sensor inputs and adaptation recipes. \\
    \midrule
    R4: Uncertainty & Repeated-run reporting~\citep{bouthillier2019unreproducible,bouthillier2021accountability}. & Repeat affordable evaluations. For costly dense or time-series runs, disclose single-run results and avoid small-gain claims. \\
    \midrule
    R5: Shared Harness & lm-evaluation-harness, HELM, and timm's evaluation scripts~\citep{gao2024lmeval,liang2023helm}. & Version tracks and refresh tests as benchmarks saturate or overlap with pretraining data~\citep{ott2022saturation,dehghani2021benchmarks}. \\
    \midrule
    R6: Data Controls & Controlled comparisons that isolate the source of gains~\citep{lipton2019troubling}. & Fix sensor, data coverage, and budget for at least one comparison. Smaller public subsets can make this affordable. \\
    \bottomrule
  \end{tabularx}
\end{minipage}\par
\endgroup

\end{document}